%% file: iclr2025_conference.tex
\documentclass{article} 
\usepackage{iclr2025_conference,times}
\usepackage{graphicx} 
\usepackage{booktabs} 
\usepackage{enumitem}

\input{math_commands.tex}

\usepackage{url}

\title{SPREG: Structured Plan Repair with Entropy-Guided Test-Time Intervention for Large Language Model Reasoning}

\author{
    Xuan Wang\textsuperscript{1}, Yu Ming\textsuperscript{1}, Xinhao Zhong\textsuperscript{1}, Xinyu Yu\textsuperscript{1}, Wenjie Wang\textsuperscript{1}, \\ Shuai Chen\textsuperscript{1}, Wei Lin\textsuperscript{1} \\[2ex]
    \textsuperscript{1}Meituan \\
    \parbox{0.8\textwidth}{\centering \tt\small \{wangxuan39, mingyu03, zhongxinhao, yuxinyu08, wangwenjie22, chenshuai31, linwei31\}@meituan.com}
}

\iclrfinalcopy 
\date{}
\begin{document}
\maketitle

\begin{abstract}
Large Language Models (LLMs) are prone to logical hallucinations and stochastic drifts during long-chain reasoning. While Classifier-Free Guidance (CFG) can improve instruction adherence, standard static implementations often cause semantic dilution and linguistic degradation. We propose \textbf{SPREG} (Structured Plan-guided Real-time Entropy Gating), a lightweight inference-time framework for surgical error rectification. SPREG employs an adaptive dual-threshold mechanism to monitor real-time entropy, identifying sudden ``entropy spikes'' as reliable indicators of logical failure. Upon detection, it triggers a dynamic repair by replacing uninformative null-priors with reference distributions synthesized from historical high-confidence states. By modulating guidance intensity according to structured reasoning stages (e.g., Action, Observation), SPREG steers the model back to a stable manifold without compromising fluency. Our experiments demonstrate significant gains, notably a \textbf{20.0\%} absolute accuracy improvement on AIME25, while effectively suppressing uncontrolled entropy drift in complex tasks.
\end{abstract}

\section{Introduction}

The paradigm of \textbf{Classifier-Free Guidance (CFG)} \cite{sanchez2024stay_guidance} has emerged as a powerful technique for enhancing instruction adherence in generative models. By extrapolating conditional logits away from an ``unconditional'' prior, CFG forces the model to prioritize task-specific constraints. Formally, given a condition $c$ and the $(t+1)$-th token, the guided log-probability is:
\begin{equation}
\begin{split}
\log \widehat{\text{P}}_\theta(y_{t+1} | \mathbf{y}_{<t+1}, c) 
&= \log \text{P}_\theta(y_{t+1} | \mathbf{y}_{<t+1}) \\
&\quad + \gamma \left( \log \text{P}_\theta(y_{t+1} | \mathbf{y}_{<t+1}, c) 
- \log \text{P}_\theta(y_{t+1} | \mathbf{y}_{<t+1}) \right)
\end{split}
\end{equation}
where $\gamma \geq 1$ governs the guidance intensity. Despite its success in image generation, applying static CFG to Large Language Models (LLMs) reveals a critical limitation: \textit{Indiscriminate Guidance}. Since standard CFG applies a constant $\gamma$ to the entire logit vector at every decoding step, it inadvertently suppresses valid semantic candidates and introduces distribution shifts that degrade linguistic fluency~\cite{sanchez2024stay_guidance,yang2025less}. This over-correction is particularly damaging in complex reasoning, where the null prior fails to account for the evolving logical dependencies of the reasoning chain.

\begin{figure}[t]
    \centering
    \includegraphics[width=\linewidth]{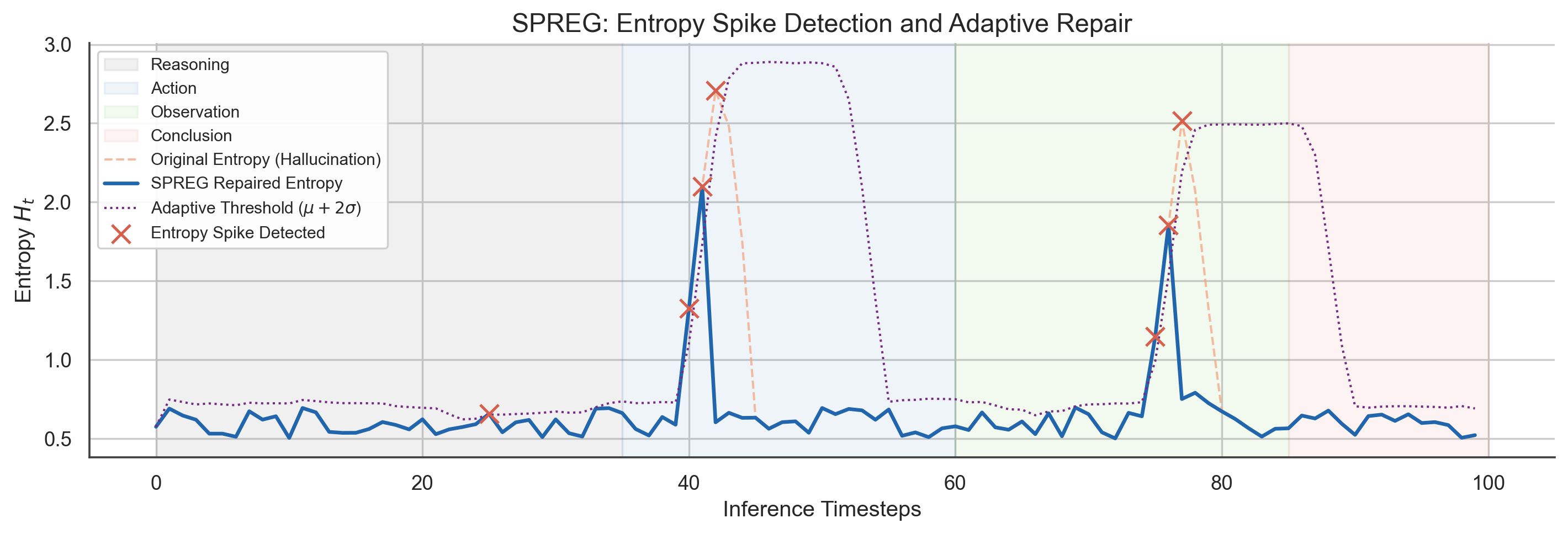}
    \caption{\textbf{Entropy Dynamics and Adaptive Repair.} The baseline entropy (dashed orange) exhibits significant spikes during logical failures within the \textit{Action} and \textit{Observation} phases. SPREG's adaptive threshold (dotted purple) identifies these anomalies (${\times}$), triggering a surgical repair that collapses the uncertainty (solid blue). This gated intervention prevents the propagation of hallucinations while maintaining the baseline's fluency during low-entropy states.}
    \label{fig:entropy_dynamics}
\end{figure}

We argue that the key to robust reasoning lies in \textbf{state-aware adaptive intervention} \cite{dong2025agentic}. Our preliminary analysis shows that the predictive entropy $H_t$ of LLMs provides a high-fidelity signal of internal hesitation. As illustrated in \textbf{Figure~\ref{fig:entropy_dynamics}}, logical inconsistencies and the onset of hallucinations are preceded by sudden ``Entropy Spikes.'' These spikes indicate that the model's autoregressive trajectory has entered a region of high epistemic uncertainty, necessitating immediate rectification.

To bridge this gap, we propose \textbf{SPREG} (\textbf{S}tructured \textbf{P}lan-guided \textbf{R}eal-time \textbf{E}ntropy \textbf{G}ating), an inference-time framework that transforms CFG from a static constraint into a dynamic repair mechanism. SPREG employs an adaptive thresholding scheme to monitor entropy fluctuations in real time, triggering intervention \emph{only} when a spike is detected. Crucially, SPREG replaces the uninformative null prior with a \textit{Dynamic Reference Distribution} derived from the model's own low-entropy historical states. By selectively sharpening the output distribution at points of failure, SPREG steers the model back toward a high-confidence reasoning path without the semantic dilution associated with traditional CFG. Our contributions are as follows:

\begin{itemize}[leftmargin=1.5em]
    \item We characterize the entropy dynamics of LLMs during structured reasoning and identify entropy spikes as a reliable leading indicator of logical hallucinations.
    \item We propose SPREG, an inference-time framework that combines plan-guided gating with an adaptive CFG mechanism for surgical error rectification.
    \item We demonstrate that SPREG significantly enhances the accuracy of reasoning on challenging benchmarks (e.g. + 20.0\% on AIME25) while preserving linguistic integrity.
\end{itemize}

\section{Related Work}

\paragraph{Classifier-Free Guidance for Language Models.}
CFG was originally proposed for diffusion-based image generation~\cite{ho2022cfg}, where it steers the denoising trajectory by amplifying the gap between conditional and unconditional score estimates. Recent work has adapted this idea to autoregressive LLMs. \citet{sanchez2024stay} showed that applying CFG at the token level can improve instruction following, while \citet{yang2025less} demonstrated that reducing guidance on high-confidence tokens mitigates over-suppression of valid continuations. Despite these advances, all existing LLM-CFG methods apply a \emph{fixed} guidance scale throughout generation, leaving them vulnerable to the indiscriminate guidance problem we identify in this work.

\paragraph{Uncertainty Estimation and Entropy in LLMs.}
Predictive entropy has long been used as a proxy for model uncertainty in Bayesian deep learning~\cite{gal2016dropout}. In the context of LLMs, \citet{kadavath2022language} showed that well-calibrated models exhibit higher entropy on questions they are likely to answer incorrectly. \citet{malinin2021uncertainty} studied ensemble-based uncertainty decomposition for sequence models, and \citet{kuhn2023semantic} proposed semantic entropy as a more robust uncertainty measure for free-form generation. Our work builds on this line of research by treating token-level entropy as a real-time signal for detecting reasoning failures, rather than as a post-hoc evaluation metric.

\paragraph{Inference-Time Scaling and Reasoning.}
A growing body of work explores how to improve LLM reasoning at inference time without modifying model weights. Chain-of-Thought (CoT) prompting~\cite{wei2022chain} elicits multi-step reasoning by conditioning on intermediate steps. Self-consistency~\cite{wang2023selfconsistency} improves reliability by sampling multiple reasoning paths and taking a majority vote. More recent approaches such as Tree-of-Thought~\cite{yao2023tree} and Monte Carlo Tree Search (MCTS)-based methods~\cite{zhang2024rest} explicitly search over reasoning trajectories. Process Reward Models (PRMs)~\cite{lightman2023lets} provide step-level supervision to guide search. SPREG is complementary to these approaches: rather than searching over multiple trajectories, it \emph{repairs} a single trajectory on the fly by detecting and correcting entropy spikes, making it compatible with standard greedy or beam-search decoding.

\paragraph{Hallucination Detection and Mitigation.}
Hallucination in LLMs has been studied extensively from both the training and inference perspectives~\cite{ji2023survey}. Retrieval-augmented generation (RAG)~\cite{lewis2020retrieval} grounds model outputs in external knowledge, while contrastive decoding~\cite{li2023contrastive} suppresses outputs that are disproportionately likely under a weaker model. \citet{chuang2023dola} proposed DoLa, which contrasts later and earlier layer logits to reduce factual errors. Closest to our work, \citet{dong2025agentic} explored agentic intervention strategies that monitor internal model states to trigger corrective actions. SPREG differs from these methods in that it uses entropy dynamics—rather than external knowledge or layer-wise contrasts—to identify the precise token positions where hallucinations are imminent, and applies a targeted CFG correction only at those positions.

\paragraph{Structured Reasoning and Plan Tracking.}
Several works have explored imposing explicit structure on LLM reasoning chains. Scratchpad methods~\cite{nye2021show} encourage models to write intermediate computations, while \citet{zhou2023leasttomost} decompose complex problems into ordered sub-problems. In the agentic setting, ReAct~\cite{yao2023react} interleaves reasoning and action steps with structured keywords. SPREG's PlanTracker component is inspired by this line of work: it identifies step boundaries (reasoning, action, observation, conclusion) via lightweight keyword matching, enabling context-aware guidance that reduces intervention during stable conclusion steps.

\section{Methodology}

\begin{figure*}[t]
    \centering
    \includegraphics[width=1.0\linewidth]{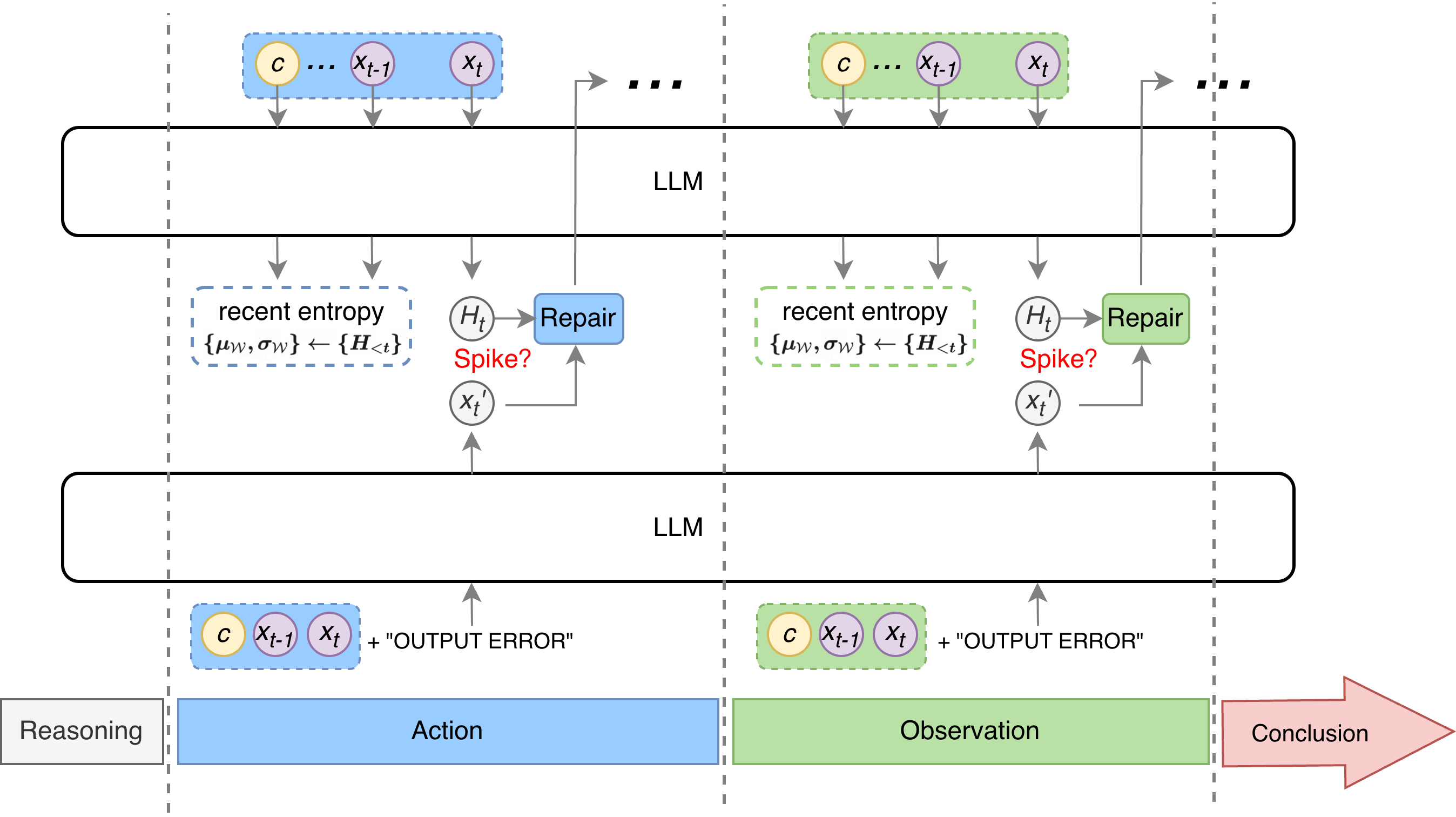}
    \caption{\textbf{Overall Execution Pipeline of the SPREG framework.} SPREG functions as a lightweight inference-time wrapper designed to enhance the reliability of LLM inference. \textbf{(Bottom)} The \textit{PlanTracker} partitions the generation into logical segments (e.g., \textit{Action}, \textit{Observation}). \textbf{(Middle)} For each decoding step $t$, the system parallely observes the base LLM output to compute the Shannon entropy $H_t$ and maintains a sliding window $\mathcal{W}$ to calculate adaptive thresholds $\{\mu_{\mathcal{W}}, \sigma_{\mathcal{W}}\}$. If an anomaly is detected ($H_t > \tau$), the \textbf{Adaptive CFG Repair} is activated to modulate the logits. This ensures precise, token-level intervention, effectively rectifying errors induced by unexpected contexts (e.g., ``OUTPUT ERROR'').}
    \label{fig:spreg_pipeline}
\end{figure*}

We introduce \textbf{SPREG} (Structured Plan-guided Real-time Entropy Gating), a lightweight framework designed to enhance the reliability of Large Language Models (LLMs) during complex reasoning. As illustrated in \textbf{Figure~\ref{fig:spreg_pipeline}}, SPREG operates via an adaptive ``Monitor-Detect-Repair'' loop, selectively intervening only when the model's predictive trajectory exhibits high-uncertainty anomalies.

\subsection{Adaptive Detection via Historical Entropy Patterns}

To identify potential reasoning failures, SPREG monitors the Shannon entropy $H_t$ of the predictive distribution $P_t$ at each decoding step $t$:
\begin{equation}
H_t = -\sum_{v \in V} p_{t,v} \log p_{t,v}
\end{equation}
Rather than relying on static thresholds, SPREG maintains a sliding window $\mathcal{W}_t$ of size $W=10$ to estimate the local manifold of model uncertainty.

\begin{table}[h]
\centering
\caption{\textbf{Semantic Patterns for Structural Step Identification.} The \textit{PlanTracker} utilizes these multi-lingual keyword patterns and syntax features to dynamically partition the autoregressive stream into functional segments.}
\label{tab:step_patterns}
\begin{tabular}{lll}
\toprule
\textbf{Step Type} & \textbf{Key Semantics} & \textbf{Syntactic / Format Cues} \\
\midrule
\textbf{Reasoning} & First, Let me, Thinking & Step-by-step markers (e.g., Step 1) \\
\textbf{Action} & Tool, Action, Function & Code blocks, JSON, \texttt{def}, \texttt{import} \\
\textbf{Observation} & Result, Output, Observation & Indicators like \texttt{=>}, \texttt{→}, API returns \\
\textbf{Conclusion} &Therefore, Final answer, Thus & \textit{Fixed tags} (e.g., The answer is) \\
\bottomrule
\end{tabular}
\end{table}

\subsubsection{Dual-Threshold Spike Detection}
To distinguish meaningful ``hesitation'' from inherent linguistic variance, we define an \textbf{Entropy Spike} using a dual-threshold mechanism. An intervention is triggered only if:
\begin{equation}
\text{is\_spike} = \mathbf{1}\!\left[H_t > \mu_W + \alpha \cdot \sigma_W\right] \wedge \mathbf{1}\!\left[H_t \geq H_{\min}\right]
\end{equation}
where $\mu_W $ and $\sigma_W$ are the moving average and standard deviation within $\mathcal{W}_t$. 
\textbf{Motivation:} The relative sensitivity $\alpha=1.5$ captures local anomalies, while the absolute floor $H_{\min}=2.0$ filters out stochastic noise in high-confidence regions, ensuring the precision of the trigger.

\subsubsection{Gradient-based Pre-filtering}
Before triggering the dual-threshold check, we analyze the \textbf{Entropy Gradient} $g_t$ over the last $n=5$ steps to estimate the instantaneous rate of uncertainty change:
\begin{equation}
g_t = \frac{\sum_{i=0}^{n-1}(i - \bar{i})(H_{t-n+1+i} - \bar{H})}{\sum_{i=0}^{n-1}(i - \bar{i})^2}
\end{equation}
where $\bar{i} = (n-1)/2$. If the entropy increases slowly ($g_t < g_{\min}=0.3$) and remains moderate ($H_t < 3.5$), it is categorized as a natural semantic transition. Only rapid surges ($g_t \geq g_{\min}$) or extreme states ($H_t \geq 3.5$) are treated as potential failures.

\subsubsection{Temporal Intervention Control}
To preserve fluency and avoid cumulative bias, SPREG employs a three-phase control rhythm:
\begin{itemize}[leftmargin=1.5em, itemsep=2pt]
    \item \textbf{Warm-up:} No detection occurs during the first $T_{\text{warm}}=5$ steps to allow the entropy window to stabilize.
    \item \textbf{Repair:} Upon a trigger, intervention persists for $d_{\text{repair}} \in \{1, 2, 3\}$ steps (dynamically decided by severity) to stabilize the reasoning path.
    \item \textbf{Cooldown:} A period of $T_{\text{cool}}=30$ steps follows each repair to prevent frequent oscillations and over-correction.
\end{itemize}

\subsection{Entropy-Guided Adaptive Plan Repair}

When a spike is detected, SPREG modulates the output distribution using an entropy-aware \textbf{Classifier-Free Guidance (CFG)} framework.

\subsubsection{Dynamic CFG Formulation}
Unlike standard CFG which utilizes a fixed prior, we define guided logits $\tilde{\mathbf{z}}_t$ that extrapolate away from a \textbf{dynamic reference distribution} $\mathbf{z}_t^{\text{ref}}$:
\begin{equation}
\tilde{\mathbf{z}}_t = \mathbf{z}_t^{\text{ref}} + \lambda(H_t, \tau, r) \cdot \mathbf{w}_t \odot \left(\mathbf{z}_t^c - \mathbf{z}_t^{\text{ref}}\right)
\end{equation}
where $\mathbf{z}_t^c$ denotes conditional logits, $\lambda$ is the adaptive guidance scale, and $\mathbf{w}_t$ is a token-level weight. For numerical stability, the computation is performed in log-softmax space: $\tilde{\ell}_t = \lambda (\ell_t^c - \ell_t^{\text{ref}}) + \ell_t^{\text{ref}}$.

\subsubsection{Reference Prior and Efficiency}
The reference prior $\mathbf{z}_t^{\text{ref}}$ represents the model's grounded state, synthesized by aggregating historical low-entropy distributions where $H_s < \mu_W$. In practice, we approximate this prior via an unconditional forward pass with a neutral prompt. By setting \texttt{dont\_save\_kv\_cache=True}, we reuse the existing KV cache of the conditional prefix without contamination, significantly reducing the computational overhead of the guidance.

\subsubsection{Adaptive Guidance Scaling and Token Weighting}
The scale $\lambda$ is dynamically scaled based on the reasoning step type $\tau$ (from \textbf{Table~\ref{tab:step_patterns}}), current entropy, and repair count $r$:
\begin{equation}
\lambda(H_t, \tau, r) = \min\!\left(\lambda_{\text{base}}(\tau) \cdot \left(1 + \beta \frac{H_t - \mu_W}{\mu_W + \epsilon}\right) \cdot \gamma(\tau) \cdot \delta(r),\ \lambda_{\max}\right)
\end{equation}
where $\delta(r) = 1/(1+r)$ is a decay factor. Furthermore, we introduce \textbf{Token-level Weighting} $\mathbf{w}_t$ based on the reference distribution variance:
\begin{equation}
\mathbf{w}_t(v) = 1 + \eta \cdot \hat{H}_t \cdot \frac{\mathbf{z}_t^{\text{ref}}(v) - \mu_{\text{ref}}}{\sigma_{\text{ref}} + \epsilon}
\end{equation}
where $\hat{H}_t = H_t / \log|V|$ is the normalized entropy. This weighting penalizes anomalous tokens that deviate from the high-confidence mean, providing surgical precision during intervention.

\subsubsection{Aggressive Recovery and Loop Detection}
In cases of persistent uncertainty (e.g., $C_{\text{high}} \geq 50$ consecutive steps above $\mu_W$), SPREG activates an \textbf{Aggressive Recovery Mode}. In this state, we amplify $\lambda$, apply a sharp temperature $T=0.3$, and impose a repetition penalty $\rho$ to break logical loops:
\begin{equation}
\hat{z}_t(v) \leftarrow 
\begin{cases} 
\hat{z}_t(v) / \rho & \text{if } v \in \mathcal{V}_{\text{recent}} \text{ and } \hat{z}_t(v) > 0 \\
\hat{z}_t(v) \cdot \rho & \text{if } v \in \mathcal{V}_{\text{recent}} \text{ and } \hat{z}_t(v) < 0 
\end{cases}
\end{equation}
where $\mathcal{V}_{\text{recent}}$ is the set of tokens generated in the previous window. This forces the model to explore new semantic paths and exit the deterministic error state.

\section{Experiments}

\subsection{Experimental Setup}
\textbf{Benchmarks.} We evaluate SPREG on a spectrum of challenging reasoning tasks: (1) \textbf{Mathematical Reasoning}: AIME24\cite{aime24}, AIME25\cite{aime25}, AMC23, MATH-500, and Minerva; (2) \textbf{Scientific Reasoning}: GPQA\cite{rein2024gpqa}; (3) \textbf{Comprehensive Benchmarks}: Olympiad-Bench\cite{he2024olympiadbench}. These datasets are selected to test the model's ability to maintain logical consistency over extended reasoning chains.

\textbf{Implementation Details.} Our experiments are conducted using Qwen3-8B\cite{yang2025qwen3} as the base model. The sliding window size for entropy monitoring is set to $\mathcal{W}=10$, and the spike sensitivity $\alpha$ is 2.0. We implement the PlanTracker using highly optimized regular expressions. For the adaptive repair, we set the base guidance scale $\gamma_{\text{base}}$ to $1.5$ for reasoning steps and $1.8$ for action/conclusion steps.

\subsection{Main Results}

\begin{table*}[htbp]
\centering
\caption{\textbf{Main Results on Reasoning Benchmarks.} We report the accuracy (Acc), percentage (\%), and the absolute improvement. \textit{Spikes} and \textit{Repairs} denote the cumulative number of detected anomalies and successful adaptive interventions, respectively.}
\label{tab:main_results}
\resizebox{\textwidth}{!}{
\begin{tabular}{l|c|cc|cc|c|cc}
\toprule
\textbf{Dataset} & \textbf{Total} & \textbf{Baseline} & \textbf{Base \%} & \textbf{SPREG} & \textbf{SPREG \%} & \textbf{Improv.} & \textbf{Spikes} & \textbf{Repairs} \\
\midrule
AIME24         & 30   & 23  & 76.7\% & 25  & 83.3\% & +6.6\% & 819    & 819    \\
AIME25         & 30   & 17  & 56.7\% & 23  & 76.7\% & +20.0\% & 962    & 962    \\
AMC23          & 40   & 39  & 97.5\% & 39  & 97.5\% & +0.0\% & 691    & 691    \\
GPQA           & 198  & 98  & 49.5\% & 96  & 48.5\% & -1.0\% & 9,456  & 9,456  \\
MATH-500       & 500  & 356 & 71.2\% & 355 & 71.0\% & -0.2\% & 4,559  & 4,559  \\
Minerva        & 272  & 79  & 29.0\% & 79  & 29.0\% & +0.0\% & 5,787  & 5,787  \\
Olympiad-Bench & 674  & 327 & 48.5\% & 331 & 49.1\% & +0.6\% & 14,948 & 14,948 \\
\midrule
\textbf{Total} & \textbf{1744} & \textbf{939} & \textbf{53.8\%} & \textbf{948} & \textbf{54.4\%} & \textbf{+0.6\%} & \textbf{37,222} & \textbf{37,222} \\
\bottomrule
\end{tabular}
}
\end{table*}

Table~\ref{tab:main_results} summarizes the performance. SPREG achieves significant gains on the hardest math competition datasets (AIME24 and AIME25), where precise logical control is paramount. Interestingly, while the overall accuracy gain on larger benchmarks like MATH-500 is marginal, the high volume of \textit{Spikes} (37,222 in total) underscores the prevalence of uncertainty during reasoning, which SPREG successfully manages without degrading fluency.

\begin{figure*}[htbp]
    \centering
    \includegraphics[width=\textwidth]{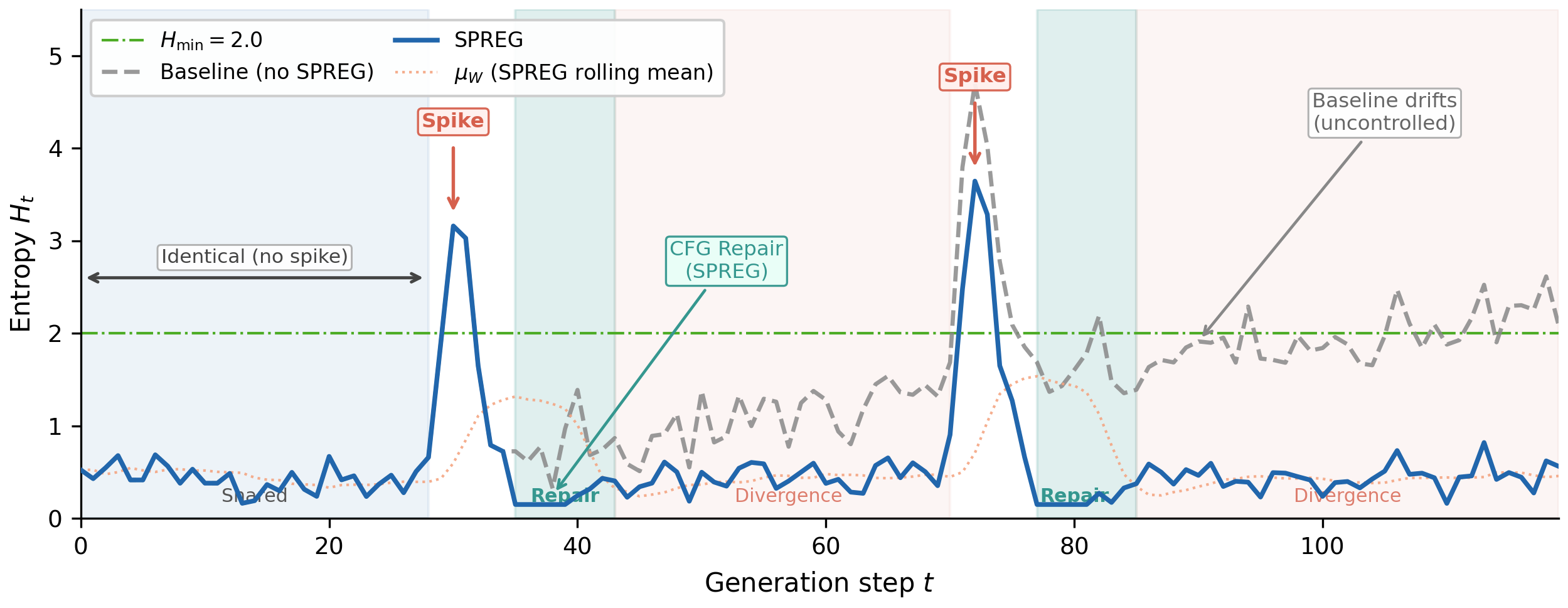}
    \caption{\textbf{Entropy Trajectory Analysis.} SPREG versus baseline during complex reasoning. \textbf{(Spike)} Adaptive detection of uncertainty surges (red arrows). \textbf{(Repair)} Entropy-Aware CFG (green regions) restores the model to a stable manifold, while the baseline exhibits uncontrolled divergence.}
    \label{fig:entropy_comparison}
\end{figure*}

\subsection{Qualitative Analysis: Entropy Dynamics and Error Rectification}

To further understand how SPREG stabilizes reasoning, we analyze the entropy trajectory
in \textbf{Figure~\ref{fig:entropy_comparison}}, yielding several key insights:

\begin{itemize}[leftmargin=1.5em, itemsep=2pt]
    \item \textbf{Proactive Anomaly Detection:} The 37,222 total spikes reported in
    \textbf{Table~\ref{tab:main_results}} highlight the ubiquity of uncertainty in
    long-chain reasoning. \textbf{Figure~\ref{fig:entropy_comparison}} shows that SPREG
    identifies critical ``decision forks'' (red arrows) before the model commits to
    erroneous tokens.

    \item \textbf{Effective Trajectory Control:} Unlike the baseline (dashed grey line)
    which exhibits uncontrolled entropy drift and divergence, SPREG's intervention (green
    regions) successfully pulls the entropy back to a stable, low-uncertainty manifold
    ($H_t < 1.0$).

    \item \textbf{Precision and Robustness:} The $H_{\min}=2.0$ floor prevents unnecessary
    intervention in fluent, low-uncertainty zones. The significant gains on
    \textbf{AIME25 (+20.0\%)} and \textbf{AIME24 (+6.6\%)} demonstrate SPREG's efficacy
    in high-stakes competition tasks where a single logical slip leads to complete failure.

    \item \textbf{Loop Suppression:} SPREG effectively mitigates the upward entropy drift
    characteristic of circular reasoning or ``rambling'' observed in the late stages of
    baseline generation, contributing to the consistent gains on Olympiad-Bench (+0.6\%).

    \item \textbf{Limitations of Confidence-Based Guidance:} A small number of datasets
    exhibit marginal performance degradation (GPQA: $-1.0\%$; MATH-500: $-0.2\%$).
    We attribute this to an inherent limitation of entropy-based intervention: SPREG
    amplifies the model's own distributional tendencies, and therefore cannot correct
    errors that arise from \emph{overconfident} mispredictions. When the model assigns
    extremely low entropy to an incorrect reasoning path---i.e., it is confidently wrong
    ---no spike is detected and no repair is triggered. Moreover, by sharpening the
    distribution at intervention points, SPREG may inadvertently suppress the stochastic
    diversity that would otherwise allow a correct answer to emerge via random sampling.
    This effect is most pronounced on datasets such as GPQA, whose questions demand
    broad world knowledge that may lie outside the model's reliable competence boundary,
    making high-confidence errors more frequent. Addressing this limitation, for instance
    by incorporating external verifiers or process reward signals to detect confident-yet-wrong
    states, is a promising direction for future work.
\end{itemize}

\section{Conclusion and Future Work}

In this paper, we introduced \textbf{SPREG}, a novel inference-time framework that enhances the logical consistency of LLMs through structured plan-guided entropy monitoring and adaptive repair. By leveraging real-time entropy dynamics and the structural priors of reasoning trajectories, SPREG effectively detects and rectifies logical drifts before they propagate into hallucinations. Our experimental results across challenging benchmarks, such as AIME24 and AIME25, demonstrate that SPREG not only improves reasoning accuracy but also maintains high generation stability with minimal computational overhead.

The success of SPREG opens several promising avenues for future research:

\begin{itemize}[leftmargin=1.5em, itemsep=2pt]
    \item \textbf{Adaptive Token-level Loss Weighting for Post-training:} A natural extension of this work is to integrate the entropy signals used in SPREG directly into the model's \textit{training phase}. Traditional Supervised Fine-Tuning (SFT) and RLHF often suffer from over-fitting due to uniform loss weighting across all tokens. Future research could explore a post-training methodology that applies differential loss weights based on token-level entropy and confidence signals \cite{dagrpo,rto_2025}. By penalizing high-uncertainty or critical reasoning tokens more heavily during optimization, we could potentially mitigate hallucination tendencies at the fundamental model level.
    \item \textbf{Cross-Model Uncertainty Knowledge Distillation:} We also aim to investigate whether the entropy patterns captured by SPREG can be distilled from larger teacher models into smaller student models. This could enable efficient, "uncertainty-aware" small models that inherit the robust decision-making boundaries of their larger counterparts.
\end{itemize}

\bibliography{iclr2025_conference}
\bibliographystyle{iclr2025_conference}

\end{document}

%% file: math_commands.tex
\usepackage{amsmath,amsfonts,bm}

\def\eqref#1{equation~\ref{#1}}

\def\1{\bm{1}}

\DeclareMathAlphabet{\mathsfit}{\encodingdefault}{\sfdefault}{m}{sl}
\SetMathAlphabet{\mathsfit}{bold}{\encodingdefault}{\sfdefault}{bx}{n}

